# Bayesian network learning by compiling to weighted MAX-SAT

James Cussens
Department of Computer Science &
York Centre for Complex Systems Analysis
University of York
Heslington, York YO10 5DD, UK
jc@cs.york.ac.uk

## Abstract

The problem of learning discrete Bayesian networks from data is encoded as a weighted MAX-SAT problem and the MaxWalkSat local search algorithm is used to address it. For each dataset, the per-variable summands of the (BDeu) marginal likelihood for different choices of parents ('family scores') are computed prior to applying MaxWalkSat. Each permissible choice of parents for each variable is encoded as a distinct propositional atom and the associated family score encoded as a 'soft' weighted single-literal clause. Two approaches to enforcing acyclicity are considered: either by encoding the ancestor relation or by attaching a total order to each graph and encoding that. The latter approach gives better results. Learning experiments have been conducted on 21 synthetic datasets sampled from 7 BNs. The largest dataset has 10,000 datapoints and 60 variables producing (for the 'ancestor' encoding) a weighted CNF input file with 19,932 atoms and 269,367 clauses. For most datasets, MaxWalkSat quickly finds BNs with higher BDeu score than the 'true' BN. The effect of adding prior information is assessed. It is further shown that Bayesian model averaging can be effected by collecting BNs generated during the search.

## 1 Introduction

Bayesian network learning is a hard combinatorial optimisation problem which motivates applying state-of-the-art algorithms for solving such problems. One of the most successful algorithms for solving SAT, the satisfiability problem in clausal propositional logic, is the local search WalkSAT algorithm [8]. The basic idea is to search for a satisfying assignment of a CNF formula by flipping the truth-values of atoms in that CNF. Both 'directed' flips which decrease the number of unsatisfied clauses and random flips are used. Frequent random restarts ('tries') are also used.

The SAT problem can be extended to the weighted MAX-SAT problem where weights are added to each clause and the goal is to find an assignment that maximises the sum of the weights of satisfied clauses (equivalently minimises the sum of the weights of unsatisfied clauses which can then be viewed as costs). WalkSAT can be extended to MaxWalkSAT where truth value flipping is a mixture of random flips and those which aim to reduce the cost of unsatisfied clauses.

Problems of *probabilistic inference in* Bayesian networks have already been encoded as weighted MAX-SAT problems and various algorithms, including MaxWalkSAT have been applied to them [6, 7]. Similar approaches combined with MCMC have been used for inference in Markov logic [5]. The current paper appears to be the first to apply a MAX-SAT encoding to *learning of* Bayesian networks.

The paper is structured as follows. Section 2 describes the synthetic datasets used. Section 3 describes how the necessary weights were extracted from these datasets. Section 4 is the key section where the method of encoding a BN learning problem as a weighted MAX-SAT one is given. Section 5 provides empirical evaluation of the technique for both model selection and model averaging. There then follow conclusions and pointers to future work in Section 6.

All runs of MaxWalkSAT were conducted using the C implementation (sometimes slightly adapted) available from the WalkSAT home page http://www.cs.rochester.edu/~kautz/walksat/. All computations were performed using a 2.40GHz Pentium running under GNU/Linux.

|  | | max | |
|---|---|---|---|
| Name | $n$ | \|Pa\| | $r$ |
| Mildew | 35 | 3 | 100 |
| Water | 32 | 5 | 4 |
| alarm | 37 | 4 | 4 |
| asia | 8 | 2 | 2 |
| carpo | 60 | 5 | 4 |
| hailfinder | 56 | 4 | 11 |
| insurance | 27 | 3 | 5 |

Table 1: The seven Bayesian networks used in this paper. $n$ is the number of variables, $\max |\text{Pa}|$ is the size of the biggest parent set and $\max r$ is the maximum number of values associated with a variable in the network.

|  | $N = \ldots$ | | |
|---|---|---|---|
| Name | $10^2$ | $10^3$ | $10^4$ |
| Mildew | 100 | 1000 | 10000 |
| Water | 100 | 990 | 9368 |
| alarm | 94 | 805 | 5970 |
| asia | 16 | 30 | 52 |
| carpo | 100 | 995 | 9662 |
| hailfinder | 100 | 1000 | 10000 |
| insurance | 99 | 982 | 8918 |

Table 2: Number of distinct joint instantiations sampled from each 'true' BN where $N$ is the total number of joint instantiations.

## 2 Bayesian networks and datasets

Since the goal of this paper is to evaluate a particular approach to BN learning, all datasets are the result of sampling from a known 'true' BN. Seven BNs were used: their names and key characteristics are given in Table 1. Each BN only involves discrete variables. All BNs were obtained in Bayesian Interchange (`.bif`) format from the Bayesian Network Repository (http://compbio.cs.huji.ac.il/Repository/). Each was then then converted into a Python class instance and 'pickled' (i.e. saved to disk).

Datasets of sizes 100, 1,000 and 10,000 were then generated by forward sampling from each BN. Note that each dataset is complete: there are no missing values. Each dataset was stored as a single table in an SQLite database. Each such table had two columns: a string representation of each distinct joint instantiation sampled and a count of how often that joint instantiation had been sampled. For big BNs most or all of these counts will be one. Table 2 states the number of distinct joint instantiations for each dataset. Each SQLite database was wrapped inside a Python class instance which was then pickled.

## 3 Computing and filtering family scores

The primary goal of this paper is to evaluate MaxWalkSat as an algorithm to search for BNs of high marginal likelihood for a given dataset. As is well known, if Dirichlet priors are used for the parameters of a BN and the data are complete then the marginal likelihood for a BN structure $G$ for data $D$ is given by the equations in (1) and (2) [3].

$$P(G|D) = L(G) = \prod_{i=1}^{n} \text{Score}_i(G) \qquad (1)$$

where

$$\text{Score}_i(G) = \prod_{j=1}^{q_i} \frac{\Gamma(\alpha_{ij})}{\Gamma(n_{ij} + \alpha_{ij})} \prod_{k=1}^{r_i} \frac{\Gamma(n_{ijk} + \alpha_{ijk})}{\Gamma(\alpha_{ijk})} \qquad (2)$$

Equation (2) defines what will be called a *family score* since it is determined by a child variable and its parents. In (2), $q_i$ is the number of joint instantiations of those parents that $X_i$ has in the graph $G$; $j$ is thus an index over these instantiations. $r_i$ is the number of values $X_i$ has and thus $k$ is an index over these values. $n_{ijk}$ is the count of how often $X_i$ takes its $k$th value when its parents are in their $j$th instantiation. $\alpha_{ijk}$ is the corresponding Dirichlet parameter. Finally, $n_{ij} = \sum_{k=1}^{r_i} n_{ijk}$ and $\alpha_{ij} = \sum_{k=1}^{r_i} \alpha_{ijk}$. Since $\text{Score}_i(G)$ only depends on $\text{Pa}_i$, the parents it has in graph $G$, $\text{Score}_i(G)$ can be written as $\text{Score}_i(\text{Pa}_i)$.

Throughout this paper the values of $\alpha_{ijk}$ have been chosen so that $L(G)$ becomes the *BDeu metric* for scoring BNs. This is done by choosing a *prior precision* value $\alpha$ and setting $\alpha_{ijk} = \alpha/(r_i q_i)$. In this paper the prior precision was always set to $\alpha = 1$. The BDeu score is a special case of a BDe score; BDe scores have the property that Markov equivalent BNs are guaranteed to have the same score. Setting $\alpha_{ijk} = 1/(r_i q_i)$ allows family scores to be defined by (3).

$$\text{Score}_i(\text{Pa}_i) = \prod_{j=1}^{q_i} \frac{\Gamma(\frac{1}{q_i})}{\Gamma(n_{ij} + \frac{1}{q_i})} \prod_{k=1}^{r_i} \frac{\Gamma(n_{ijk} + \frac{1}{r_i q_i})}{\Gamma(\frac{1}{r_i q_i})} \qquad (3)$$

Now, let $\mathbf{X}$ be any subset of the variables and define, for fixed data, the function $H$, as in (4).

$$H(\mathbf{X}) = \prod_{\ell=1}^{q_{\mathbf{X}}} \frac{\Gamma(n_\ell + \frac{1}{q_{\mathbf{X}}})}{\Gamma(\frac{1}{q_{\mathbf{X}}})} \qquad (4)$$

where $q_{\mathbf{X}}$ is the number of joint instantiations of variables $\mathbf{X}$, and $n_\ell$ is a count of how often the $\ell$th of these occurred in the data. Let $\text{Pa}_i$ be the parents of $X_i$ and set $\text{Fa}_i = \{X_i\} \cup \text{Pa}_i$ (the *family* for $X_i$), then it is easy to see that (3) can be rewritten as (5)

$$\text{Score}_i(\text{Pa}_i) = \frac{H(\text{Fa}_i)}{H(\text{Pa}_i)} \qquad (5)$$

so that

$$\log \text{Score}_i(\text{Pa}_i) = \log H(\text{Fa}_i) - \log H(\text{Pa}_i) \qquad (6)$$

|         |         | Time taken for $N = \ldots$ | | |
|---------|---------|--------|--------|--------|
| Name    | \|scores\| | $10^2$ | $10^3$ | $10^4$ |
| Mildew  | 230,300 | 1,678  | 1,942  | 4,502  |
| Water   | 159,744 | 22     | 123    | 1,093  |
| alarm   | 288,859 | 38     | 208    | 1,501  |
| asia    | 512     | 0      | 0      | 0      |
| carpo   | 2,056,800 | 544  | 1,811  | 13,892 |
| hailfinder | 1,555,456 | 852 | 2,974 | 22,666 |
| insurance | 79,704 | 29    | 97     | 749    |

Table 3: Computing all family scores with at most 3 parents. Times are in seconds rounded (asia did take a small positive amount of time!)

In the approach followed in this paper the first stage of BN learning is to compute log family scores $\log \text{Score}_i(G)$ for all families up to some limit on the number of parents. Here this limit was set to 3 parents. Note that 4 of the 7 true BNs have variables with more than 3 parents so that this restriction renders these 4 unlearnable. Equation (6) permits a small efficiency increase: $\log H(\text{Fa}_i(\mathbf{X}))$ is computed for all subsets of variables from size 0 to size 4. Family scores can then be computed using (6). (All scores will be log scores from now on.)

The number of family scores computed and the time taken is listed for each dataset in Table 3. These computations were performed using Python equipped with the Psyco (http://psyco.sourceforge.net/) JIT compiler. Re-implementing in, say, C would no doubt be faster, but optimising this part of the process is not the focus of this paper.

Once family scores are computed the next step is to encode them as weighted clauses, but, with the exception of asia, there are too many scores for all to be used. However, since the goal is to find high likelihood BNs the vast majority of these family scores can be thrown away. Let $\text{Pa}_i$ and $\text{Pa}'_i$ be two potential parent sets for variable $X_i$. If (1) $\text{Pa}_i \subset \text{Pa}'_i$ and (2) $\text{Pa}_i$ has a higher score than $\text{Pa}'_i$ then $\text{Pa}'_i$ can be ruled out as a potential parent set for $X_i$ in a maximal likelihood graph. Any graph which has $\text{Pa}'_i$ as the parents for $X_i$ can have its score increased by replacing $\text{Pa}'_i$ by $\text{Pa}_i$; the key point is that such a replacement cannot introduce a directed cycle since one or more arrows are being removed. Filtering family scores in this way causes a significant reduction in the potential parent sets for any variable as shown by Table 4.

|     | $N = \ldots$ | | | | | |
|-----|-----|-----|-----|-----|-----|-----|
|     | $10^2$ | | $10^3$ | | $10^4$ | |
| Dat | max | $n$ | max | $n$ | max | $n$ |
| Mi  | 977 | 3,515 | 17  | 163 | 47  | 465 |
| Wa  | 44  | 482 | 44  | 573 | 49  | 961 |
| al  | 75  | 907 | 184 | 1,928 | 473 | 6,473 |
| as  | 10  | 41  | 24  | 107 | 31  | 161 |
| ca  | 350 | 5,068 | 352 | 3,827 | 2,089 | 16,391 |
| ha  | 22  | 244 | 77  | 761 | 435 | 3,768 |
| in  | 28  | 279 | 95  | 774 | 496 | 3,652 |

Table 4: Results of filtering potential parents for variables. 'True' BN names have been abbreviated. $n$ is the total number of candidate parent sets for all variables. max is the number of candidate parents for that variable which has the most candidate parents.

## 4  Encoding BN learning using hard and soft clauses

The goal of finding a maximum (marginal) likelihood BN with complete data is equivalent to choosing the $n$ best parent sets (one for each variable) subject to the condition that no directed cycle is formed: a problem of combinatorial optimisation. Here this problem is encoded with weighted clauses.

A weighted CNF problem requires specifying (i) propositional atoms (ii) clauses and (iii) weights for clauses. To represent graph structure two options have been considered. Using an adjacency matrix approach for each of the $n(n-1)$ distinct ordered pairs of vertices there is an atom asserting the existence of an arc between the two vertices. Alternatively one can create one atom for each possible family (child plus parents). This latter approach has performed better in the experiments done so far, presumably because flipping the truth values of such atoms permits bigger search moves. With this approach when encoding the problem of learning from the carpo-generated dataset with 10,000 datapoints there are 16,391 such atoms (see Table 4). If such an atom is set to TRUE, this represents that the specified variable has the specified parents. For each variable there is a single clause stating that it must have a (possibly empty) parent set:

$$\bigvee_{\text{candidate parent set } t} X_i \text{ has parent set } t \qquad (7)$$

The longest such clause contains 2,089 atoms.

To rule out choices of parent sets producing a cycle two options have been considered. In the first 'Ancestor' encoding, there is an atom an$(X_i, X_j)$ stating that $X_i$ is an ancestor of $X_j$ in the graph for each of the $n(n-1)$ distinct ordered pairs of vertices. For each of the $n(n-1)(n-2)$ distinct ordered triples $(X_i, X_j, X_k)$ there is a transitivity clause:

$$\text{an}(X_i, X_j) \wedge \text{an}(X_j, X_k) \rightarrow \text{an}(X_i, X_k) \qquad (8)$$

To express acyclicity there are $n(n-1)/2$ clauses of the form:

$$\neg\text{an}(X_i, X_j) \vee \neg\text{an}(X_j, X_i) \qquad (9)$$

An alternative 'Total order' approach is to encode a total order of vertices. In a total order exactly one of $X_i < X_j$ and $X_j < X_i$ is true for each distinct $X_i$ and $X_j$. So for each of the $n(n-1)/2$ distinct unordered pairs $\{X_i, X_j\}$ there is an atom ord$(X_i, X_j)$ asserting that $X_i$ and $X_j$ are lexicographically ordered in the total order. To ensure that an assignment of truth values to these atoms defines a total order it is enough to rule out 3-cycles such as $X_i < X_j$, $X_j < X_k$, $X_k < X_i$. This is because if a fully connected digraph with no 2-cycles (a *tournament*) has a cycle then it has a 3-cycle. There is a simple inductive proof of this result which is omitted here. There are $n(n-1)(n-2)$ hard clauses ruling out 3-cycles of the form:

$$\neg\text{ord}(X_i, X_j) \vee \neg\text{ord}(X_j, X_k) \vee \text{ord}(X_i, X_k)$$
$$\text{ord}(X_i, X_j) \vee \text{ord}(X_j, X_k) \vee \neg\text{ord}(X_i, X_k)$$

Using either the 'Ancestor' or 'Total Order' encoding there are hard clauses declaring that a (non-empty) choice of parents for a variable determines the truth value of certain atoms. For example, using the 'Ancestor' approach:

$$X_j \text{ has parent set } \{X_i, X_k\} \rightarrow \text{an}(X_i, X_j)$$
$$X_j \text{ has parent set } \{X_i, X_k\} \rightarrow \text{an}(X_k, X_j)$$

or with the 'Total Order' approach:

$$X_j \text{ has parent set } \{X_i, X_k\} \rightarrow \text{ord}(X_i, X_j)$$
$$X_j \text{ has parent set } \{X_i, X_k\} \rightarrow \neg\text{ord}(X_j, X_k)$$

Note that since the log BDeu score is a log-likelihood it is a negative number, as are all the family scores. The goal is to find an admissible choice of families with small negative numbers. Another way of looking at this is to say that choosing any particular set of parents for a variable incurs a cost which is -1 times the relevant family score. Each such cost can then be encoded by a weighted clause of the following type:

$$-\log \text{Score}_i(\text{Pa}_i) : \neg(X_i \text{ has parent set } \text{Pa}_i) \qquad (10)$$

The $-\log \text{Score}_i(\text{Pa}_i)$ costs were rounded to integers.

A final option is not to rule out cycles directly, but to create a cycle_atom asserting the existence of a cycle. Cycles can then be ruled out entirely with the hard clause ¬cycle_atom or merely discouraged with a soft version. If the latter option is taken then cyclic digraphs have to be filtered out after the search is over.

The numbers of atoms, clauses and literals for an encoding of each learning problem using a cycle_atom and the 'Ancestor' encoding is given in Table 5. Without a cycle_atom there is one fewer atom and clause and also a reduction in the number of literals. With the 'Total Order' approach there is a small reduction in atoms and the number of clauses and lits is roughly halved. (A literal is an atom or its negation, the number of literals in Table 5 is the sum of all literals in all clauses.)

To help motivate these choices of encoding it is worth peeking ahead to examine a run of MaxWalkSat with the 'Ancestor' encoding as shown in Fig 1. This is a run using the encoded form of the hailfinder-generated set of 10,000 datapoints. This is a run *without* using the cycle_atom so the numbers of atoms, clauses and literals are reduced a little from those given in Table 5. The goal was to find a BN with BDeu score at least as high as that of the 'true' BN's score of -503,040 (i.e. cost of 503,400). This was achieved in just under 13 million flips taking 75 seconds.

| Data | atoms | clauses | lits |
|---|---|---|---|
| Mi_2 | 4,706 | 54,367 | 149,123 |
| Mi_3 | 1,354 | 40,807 | 122,003 |
| Mi_4 | 1,656 | 41,579 | 123,547 |
| Wa_2 | 1,475 | 32,053 | 94,793 |
| Wa_3 | 1,566 | 32,194 | 95,075 |
| Wa_4 | 1,954 | 33,352 | 97,391 |
| al_2 | 2,240 | 50,739 | 149,355 |
| al_3 | 3,261 | 53,945 | 155,767 |
| al_4 | 7,806 | 70,610 | 189,097 |
| as_2 | 98 | 488 | 1,351 |
| as_3 | 164 | 693 | 1,761 |
| as_4 | 218 | 873 | 2,121 |
| ca_2 | 8,609 | 226,406 | 661,551 |
| ca_3 | 7,368 | 221,365 | 651,469 |
| ca_4 | 19,932 | 269,367 | 747,473 |
| ha_2 | 3,325 | 170,009 | 509,305 |
| ha_3 | 3,842 | 171,400 | 512,087 |
| ha_4 | 6,849 | 181,545 | 532,377 |
| in_2 | 982 | 18,926 | 56,049 |
| in_3 | 1,477 | 20,346 | 58,889 |
| in_4 | 4,355 | 30,344 | 78,885 |

Table 5: Data on the weighted CNF provided as input to MaxWalkSat using the 'Ancestor' encoding and a cycle_atom. Datasets have been abbreviated, so that e.g. in_4 is the data set of $10^4$ datapoints sampled from **in**surance.

A key point is that the number of unsatisfied clauses in the lowest cost assignment in each try is 56, the number of variables in hailfinder. These unsatisfied clauses are 56 weighted single-literal clauses of the form (10) corresponding to a choice of parents for each variable which does not produce a cycle. In all runs on all datasets, the unsatisfied clauses in low cost assignments are *always* of this form. Choices of parents that do cause a cycle will break at least one hard clause thus incurring a sufficiently large cost that MaxWalkSat will never return them as a lowest cost assignment. Note that although only one choice of parents is possible for any variable there is no need to encode this (hard) constraint. Choosing e.g. two parent sets will always incur a higher cost then either or the two alone so the search avoids such assignments. MaxWalkSat 'wants' to avoid choosing any parents but is compelled to do so by hard constraints of the form (7).

## 5 Results

### 5.1 Learning a single BN

In this section, results using MaxWalkSat to search for a high BDeu-scoring BN are given. The scores from these searches are given in Table 6 and some timings are given in Table 7. Note that the times reported in Table 7 are for one of the slowest approaches: 'Ancestor' encoding, 50% noise (random flips). An example of such a 'slow' run was shown in Fig 1 where only 171,206 flips/sec were achieved. In contrast, using the 'Total Order' encoding with 10% noise on the same data a rate of 1,092,243 flips/secs is achievable.

One problem with evaluating a search for a high scoring BN is that the optimal BN and its score is unknown (although presumably it could be computed for the small asia learning problem). However, by choosing to evaluate on synthetic data there is at least the 'true' BN which can be scored. It is also possible to construct a high-scoring BN which meets the restriction of having at most 3 parents. Starting with the true BN replace each true parent set with the highest-scoring pre-scored parent set which is a subset of the true parent set. With the exception of carpo this produces a BN with a higher score than the true BN. The score of the BN thus constructed is given in the 'Target' column of Table 6.

|      | tries=10 |          | tries=100 |          |
| ---- | -------- | -------- | --------- | -------- |
| Data | $t$      | flip/sec | $t$       | flip/sec |
| Mi_2 | 7        | 140,656  | 68        | 146,222  |
| Mi_3 | 3        | 328,608  | 29        | 342,019  |
| Mi_4 | 5        | 193,561  | 50        | 196,502  |
| Wa_2 | 3        | 255,118  | 39        | 255,749  |
| Wa_3 | 4        | 215,222  | 45        | 218,069  |
| Wa_4 | 6        | 165,027  | 59        | 168,977  |
| al_2 | 7        | 126,217  | 79        | 125,894  |
| al_3 | 8        | 113,729  | 88        | 112,959  |
| al_4 | 10       | 92,570   | 106       | 94,280   |
| as_2 | 0        | 1,923,202| 5         | 1,977,716|
| as_3 | 0        | 1,676,086| 6         | 1,533,842|
| as_4 | 0        | 1,546,493| 6         | 1,548,088|
| ca_2 | 17       | 55,963   | 175       | 56,821   |
| ca_3 | 16       | 60,548   | 168       | 59,500   |
| ca_4 | 23       | 43,230   | 236       | 42,287   |
| ha_2 | 8        | 124,283  | 79        | 126,143  |
| ha_3 | 14       | 71,195   | 141       | 70,438   |
| ha_4 | 15       | 65,938   | 154       | 64,905   |
| in_2 | 2        | 476,978  | 20        | 498,619  |
| in_3 | 2        | 371,311  | 27        | 359,390  |
| in_4 | 4        | 217,090  | 45        | 217,800  |

Table 7: Timings corresponding to 'Ancestor' encoding runs with a 'cycle_atom' and noise at 50%. $t$ is the total time taken in seconds (rounded). Flip/sec is the frequency of literal flipping performed by MaxWalkSat.

Table 6 contains only the most interesting of many experimental runs conducted. The results show that for small datasets (*_2, corresponding to $10^2 = 100$), MaxWalkSat easily finds BNs exceeding both the True and 'Target' BN score. This reflects the relatively smooth nature of the likelihood surface. For bigger datasets, were the landscape is much more jagged, the picture is more mixed. For $10^3$ size datasets the True BN score is beaten by the 'Long' run in all cases, except for alarm. For $10^4$, the 'Long' run fails to beat carpo and insurance also. Comparing 'Ancestor' to 'Total Order' results show that the latter is the more successful encoding. Experiments unreported here have indicated that the 'Total Order' encoding works better with low noise (random flip) values and a prevention of random breaking of hard clauses. Combining these with an increased search on each try leads to the superior results in the 'Long' column of Table 6. Note that even the longest 'Long' run, ca_4, only took 32 minutes (with 510,397 flips/sec); most where much faster.

Note that in some cases MaxWalkSat returns a BN

```
newmaxwalksat version 20 (Huge)
seed = 99955222
cutoff = 10000000
tries = 100
numsol = 1
targetcost = 503040
heuristic = best, noise 50 / 100, init initfile
allocating memory...
clauses contain explicit costs
numatom = 6848, numclause = 181544, numliterals = 529296
wff read in
                                    average             average       mean
    lowest      worst    number        when                over      flips
      cost     clause    #unsat          #flips    model   success       all       until
  this try   this try  this try    this try   found      rate     tries     assign
    506076      16968        56    10000000       *          0         *          *
    501973      23318        56     2913803  2913803        50  12913803  12913803.0

total elapsed seconds = 75.428415
average flips per second = 171206
number of solutions found = 1
mean flips until assign = 12913803.000000
mean seconds until assign = 75.428415
mean restarts until assign = 2.000000
ASSIGNMENT ACHIEVING TARGET 503040 FOUND
```

Figure 1: A successful search (using the 'Ancestor' encoding) for a BN exceeding the BDeu score for the hailfinder BN using data of 10,000 data points sampled from that BN.

| Data | True | Target | Ancestor | Total order | Long | > True |
|---|---|---|---|---|---|---|
| Mi_2 | -7,786 | -6,611 | -5,711 | -5,708 | -5,705 | Y |
| Mi_3 | -63,837 | -52,866 | -47,229 | -47,194 | -47,120 | Y |
| Mi_4 | -470,215 | -459,874 | -409,641 | -410,159 | -408,282 | Y |
| Wa_2 | -1,801 | -1,523 | -1,488 | -1,486 | -1,484 | Y |
| Wa_3 | -13,843 | -13,304 | -13,293 | -13,284 | -13,247 | Y |
| Wa_4 | -129,655 | -128,745 | -129,274 | -128,916 | -128,812 | Y |
| al_2 | -1,410 | -1,383 | -1,368 | -1,368 | -1,336 | Y |
| al_3 | -11,305 | -11,255 | -11,599 | -11,501 | -11,339 | N |
| al_4 | -105,303 | -105,226 | -107,205 | -106,503 | -105,907 | N |
| as_2 | -247 | -245 | -241 | -241 | -241 | Y |
| as_3 | -2,318 | -2,318 | -2,312 | -2,312 | -2,312 | Y |
| as_4 | -22,466 | -22,466 | -22,462 | -22,462 | -22,462 | Y |
| ca_2 | -1,969 | -1,952 | -1,849 | -1,852 | -1,824 | Y |
| ca_3 | -17,739 | -17,821 | -17,938 | -17,891 | -17,731 | Y |
| ca_4 | -173,682 | -175,349 | -175,832 | -176,456 | -174,605 | N |
| ha_2 | -6,856 | -6,058 | -5,998 | -5,998 | -5,991 | Y |
| ha_3 | -55,366 | -52,747 | -53,059 | -52,881 | -52,517 | Y |
| ha_4 | -503,040 | -498,163 | -506,219 | -503,420 | -498,739 | Y |
| in_2 | -1,951 | -1,715 | -1,690 | -1,689 | -1,674 | Y |
| in_3 | -14,411 | -13,919 | -14,105 | -14,121 | -13,934 | Y |
| in_4 | -133,489 | -132,968 | -134,378 | -134,730 | -133,841 | N |

Table 6: Scores for true and target BNs (first 2 columns) followed by scores achieved using the Ancestor and Total Order encodings with 50% noise and 100,000 flips per try. 'Long' indicates using the Total Order encoding and runs using 10, 000, 000 flips per try, noise at 10% and disallowing the breaking of hard clauses by random flips. In all 3 cases 100 tries (restarts) were done. The final column indicates whether the 'Long' run produced a BN with a score exceeding that of the true BN.

which is astronomically more probable (assuming a uniform prior) than the 'true' one. For example, with Mi_2 and the Ancestor encoding, the returned BN is $e^{(-470,215+409,641)} = e^{60,574}$ more probable than the true one!

A rough comparison with Castelo and Kočka's RCARNR and RCARR algorithms [1] is possible for the al_2 and al_4 datasets. As here, datasets of size 1,000 and 10,000 were sampled from the Alarm BN and best scores of around -11,115 and -108,437 respectively were reported. This beats (resp. does not beat) our best score for Alarm, but since the actual sampled datasets are different not too much should be read into this comparison.

Some initial experiments adding prior knowledge to the 'Ancestor' encoding proved interesting. Adding in information specifying *all* known pairs of independent variables (by banning the relevant ancestor relations) had very little effect on the scores achieved. On the other hand, setting just a small number of known ancestor relations improved scores considerably. This is presumably due to the search having a bias towards sparser (less constrained) BNs.

### 5.2 Bayesian model averaging by search

Markov chain Monte Carlo (MCMC) approaches are popular for Bayesian model averaging (BMA) for Bayesian networks. The idea is to construct a Markov chain whose stationary distribution is the true posterior over BNs. By running the chain for a 'sufficient' number of iterations (and throwing away an initial burn-in) the hope is that the sample thus produced supplies a reasonable approximation to the true distribution.

Success depends on a realisation of the chain wandering into areas of high probability and visiting different BNs with a relative frequency close to the BNs' posterior probabilities. A search-based approach offers a much more direct method of BMA. The key idea is to explicitly search for areas of high probability. If, as is the case here, the probability of any visited BN can be computed at least up to a normalising constant then this value is recorded. An estimate of the posterior distribution is then found by simply assigning zero probability to any BN not encountered in the search and performing the obvious renormalisation on BN scores. An early paper proposing search for BMA is [4].

Here a crude search-based approach using MaxWalkSat and the 'Ancestor' encoding has been tried out. MaxWalkSat is run as normal and for *every* assignment where no hard clause is broken a record of the $n$ parent sets given by that assignment together with the cost of the assignment is sent to standard output. Piping this through UNIX's `sort --unique` is enough to produce a list of distinct (encoded) BNs sorted by score.

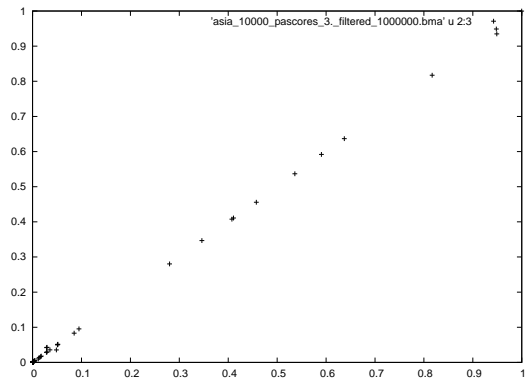

Figure 2: Comparing estimated posterior probabilities from two different BMA runs using MaxWalkSAT using asia 10,000 data

The distribution thus created can be used to estimate true posterior quantities. In experiments conducted so far only the posterior probabilities of parent sets have been estimated. By comparing estimates produced from two runs, some measure of success can be produced. Here each run used the 'Ancestor' encoding and came from 120 restarts of MaxWalkSat each using 100,000 flips. Only the 1,000,000 highest scoring BNs from each run were kept.

For the datasets considered here only asia (all sizes) and Water with 100 datapoints afforded successful BMA. Plots comparing estimated posterior probabilities for parent sets for these two cases are given in Figs 2 and 3. In all other cases such plots show big differences in estimated probabilities for most cases where the estimates differ significantly from zero. Fig 4 shows a typical excerpt of such a failure using the alarm 100 data.

The basic problem here is that, apart from the trivial asia problem, the highest scoring one or two BNs found on any run are often vastly more probable than any others, so that the estimate of the posterior distribution puts almost all the probability mass on one, occasionally two, BN(s). If these few BNs vary between runs, estimates produced from them generally vary also. Fig 5 shows a typical example of the top BN in a run being much more probable than the second-ranking one. The central issue is that often the true posterior probability really is concentrated on a few BNs, so BMA either needs to reliably find those few

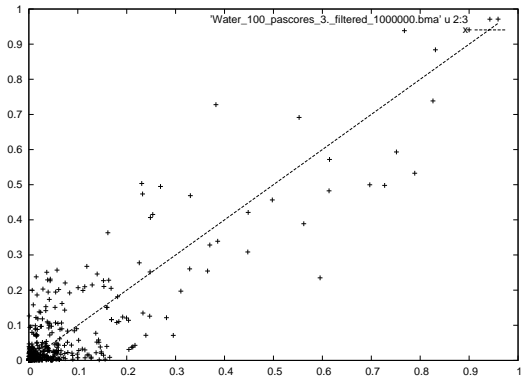

Figure 3: Comparing estimated posterior probabilities from two different BMA runs using MaxWalkSAT using Water 100 data

```
1622 9.89412087285e-35  5.71801239082e-28
1793 0.0013149823888    0.465138403992
1792 1.51462961172e-20  5.38296345824e-18
1791 5.89129801134e-14  2.10283310245e-06
1398 0.924408916903     0.00117720289971
1797 1.46223426395e-06  1.77652663003e-11
```

Figure 4: Estimates of posterior probability of parent sets from two different BMA runs for Alarm $N = 100$ dataset. The first column is the integer identifier for the parent set used by MaxWalkSat.

models or needs to move in a more regular space, such as the space of variable orderings [2].

## 6 Conclusions

The results given here should be seen as a preliminary exploration of the worth of weighted MAX-SAT encodings of BN learning and MaxWalkSat as a method for solving such encoded problems. In particular, future work needs to consider alternative encodings and greater incorporation of prior knowledge on structures. Exponential-family structure priors whose features can be easily encoded are an obvious choice; they can be encoded by weighted clauses, just as the family scores are. This paper has focused on BN learning as a pure BDeu score optimisation problem. Further work will consider the value of the learned BNs according to other metrics and provide a comparison of this learning approach with existing BN learning systems.

**How to reproduce these results** Please go to: http://www.cs.york.ac.uk/~jc/research/uai08/

## Acknowledgements

Thanks to those responsible for hosting the Bayesian Network Repository and to Henry Kautz for making MaxWalkSat available. Thanks also to 3 anonymous reviewers.

```
C 176696   3954 4048 4197 4262    ... 17333 ...
C 176714   3954 4048 4197 4262    ... 17332 ...
```

Figure 5: Top two BNs from a BMA run using the carpo $N = 10,000$ data. First number is the (rounded, negated) BDeu score. First one is 65 million times more probable (assuming a uniform prior) than second even though they differ in only one parent set (17333 vs. 17332).